%% file: egpaper_final.tex
\documentclass[10pt,twocolumn,letterpaper]{article}

\usepackage{cvpr}
\usepackage{times}
\usepackage{epsfig}
\usepackage{url}
\usepackage{color}
\usepackage{graphicx}
\usepackage{amsmath}
\usepackage{amssymb}
\usepackage{booktabs}      
\usepackage[margin=1in]{geometry}
\usepackage{array}
\usepackage{multirow}

\usepackage[breaklinks=true,bookmarks=false]{hyperref}

\cvprfinalcopy 

\def\cvprPaperID{****} 

\begin{document}

\title{Sim2Real View Invariant Visual Servoing by Recurrent Control}

\author{Fereshteh Sadeghi$^{\dag}$\thanks{Work was done while the author was an intern at Google Brain. } \hspace{1.2cm} 
Alexander Toshev$^{\ddag}$\hspace{1.2cm} 
Eric Jang$^{\ddag}$\hspace{1.2cm} 
Sergey Levine$^{\ddag}$\hspace{1.2cm} \\
\\
\hspace{-1cm}$^\dag$University of Washington, $^\ddag$ Google Brain\hspace{10cm}\\
}

\maketitle

\begin{abstract}
\input{abstract.tex}
\end{abstract}

\section{Introduction}
\input{intro.tex}

\section{Related Work}
\input{related.tex}

\section{Recurrent Models for Visual Servoing with Novel Viewpoints}
\input{approach.tex}

\section{Experiments}\label{sec:experiments}
\input{exp.tex}

\section{Discussion and Future Work}
\input{future.tex}

\section*{Acknowledgement}
We thank Erwin Coumans and Yunfei Bai for providing pybullet and Vincent Vanhoucke for helpful discussions.

{\small
\bibliographystyle{ieee}
\bibliography{egpaper_final}
}

\end{document}

%% file: abstract.tex
Humans are remarkably proficient at controlling their limbs and tools from a wide range of viewpoints and angles, even in the presence of optical distortions. In robotics, this ability is referred to as visual servoing: moving a tool or end-point to a desired location using primarily visual feedback. In this paper, we study how viewpoint-invariant visual servoing skills can be learned automatically in a robotic manipulation scenario. To this end, we train a deep recurrent controller that can automatically determine which actions move the end-point of a robotic arm to a desired object. The problem that must be solved by this controller is fundamentally ambiguous: under severe variation in viewpoint, it may be impossible to determine the actions in a single feedforward operation. Instead, our visual servoing system must use its memory of past movements to understand how the actions affect the robot motion from the current viewpoint, correcting mistakes and gradually moving closer to the target. This ability is in stark contrast to most visual servoing methods, which either assume known dynamics or require a calibration phase. We show how we can learn this recurrent controller using simulated data and a reinforcement learning objective. We then describe how the resulting model can be transferred to a real-world robot by disentangling perception from control and only adapting the visual layers. The adapted model can servo to previously unseen objects from novel viewpoints on a real-world Kuka IIWA robotic arm. For supplementary videos, see: \href{https://fsadeghi.github.io/Sim2RealViewInvariantServo}{https://fsadeghi.github.io/Sim2RealViewInvariantServo}

%% file: intro.tex
\begin{figure}[t]
\begin{center}
\includegraphics[width=\linewidth]{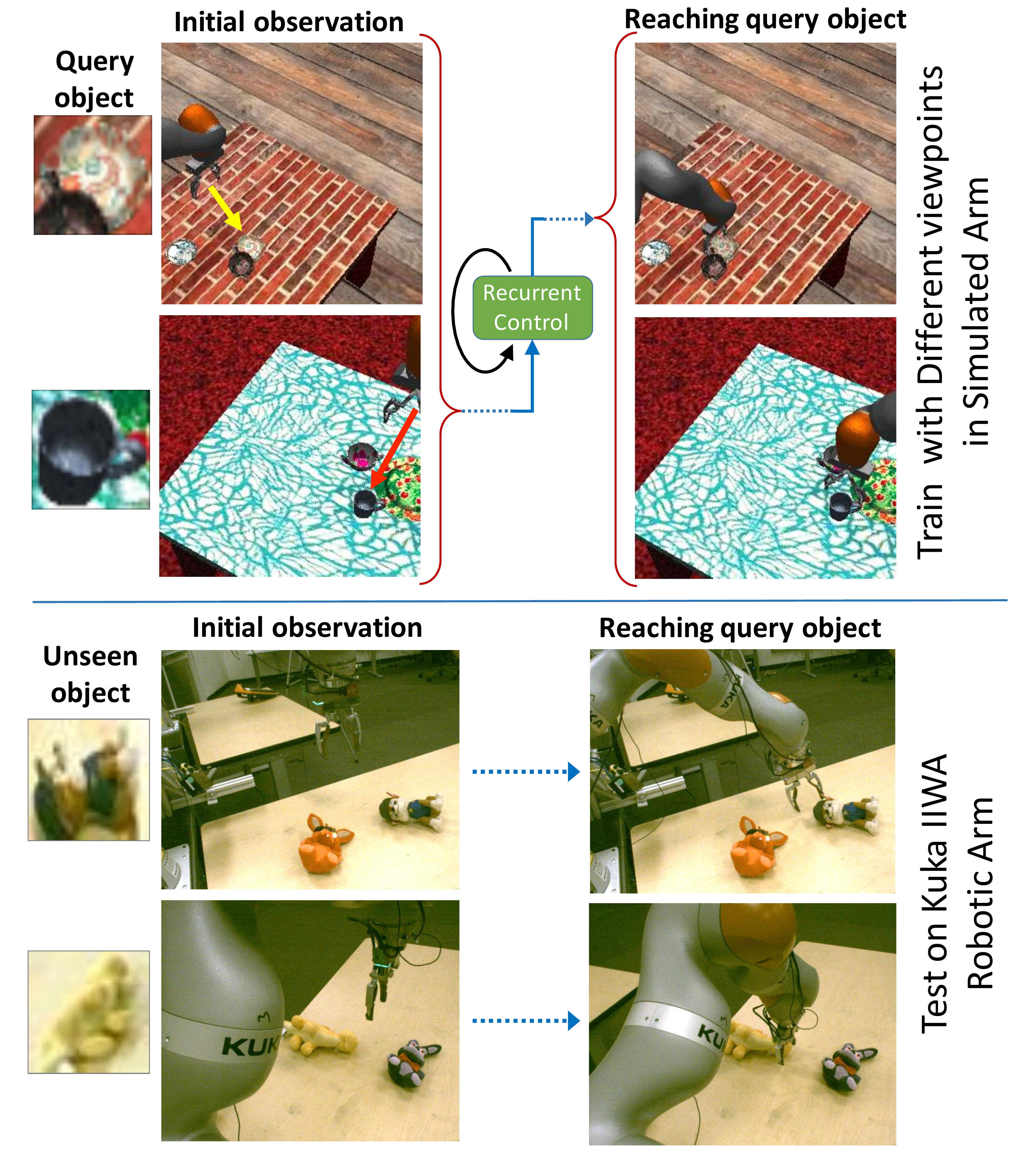}
\end{center}
\vspace{-0.2in}
\caption{Illustration of our learned recurrent visual servoing controller. Training is performed in simulation (top) to reach varied objects from various viewpoints. The recurrent controller learns to implicitly calibrate the image-space motion of the arm with respect to the actions, which are issued in the unknown coordinate frame of the robot. The model is then transferred to the real world by adapting the visual features, and can reach previously unseen objects from novel viewpoints (bottom). Depending on viewpoint, the same actions can move the arm in opposite directions, requiring the model to maintain a memory of past motions to do self-calibration and complete the task.}
\label{fig:teaser}
\vspace{-0.15in}
\end{figure}

\begin{figure*}[t]
\begin{center}
\includegraphics[width=\linewidth]{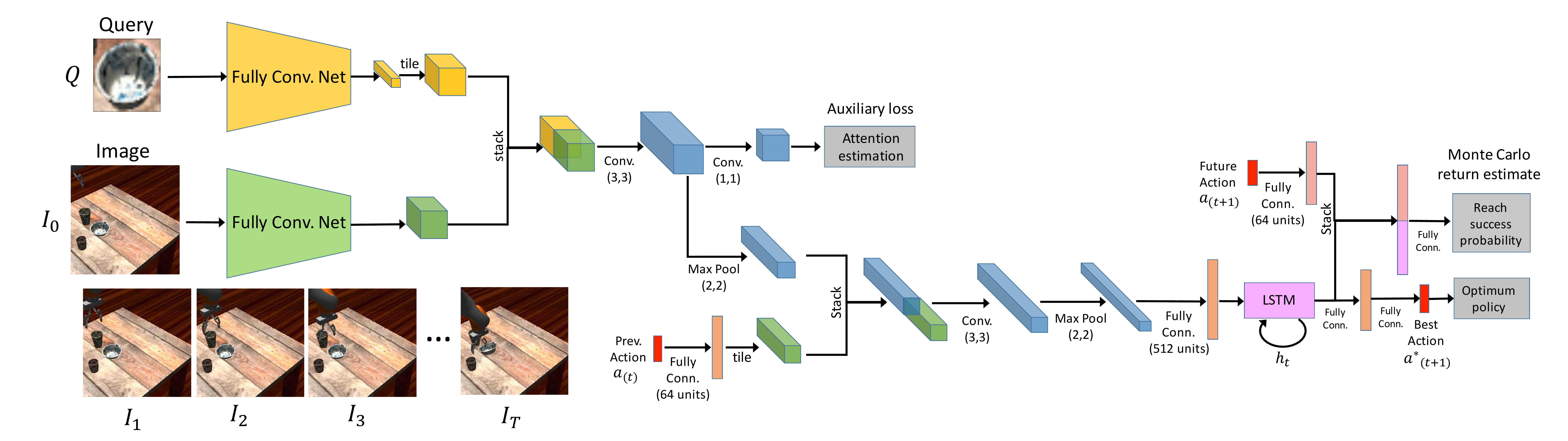}
\end{center}
\vspace{-.2in}
   \caption{\textbf{Network Architecture:} The input to the network consists of a query image (top-left) and the observed image at step $t$ (left). The images are processed by separate convolutional stacks, and their features are concatenated. The concatenated feature vector is fed into an LSTM layer and outputs the policy which is an end-effector movement command in Cartesian space, in the frame of the robot (bottom right). The previously selected action is also provided to the LSTM (bottom), enabling it to implicitly calibrate the effects of actions on image-space motion. \textbf{Value prediction:} a separate head (top right) predicts the Q-value of the action $a_t$, and is trained with Monte Carlo return estimates. \textbf{Auxiliary loss:} An auxiliary loss function minimizes the localization error for the query object in the observed image. Also used in order to adapt the convolutional layers (left) with a small number of labeled real-world images.}
\vspace{-.15in}
\label{fig:architecture}
\end{figure*}

Humans and animals can quickly recognize the effects of their actions through visual perception: when we see ourselves in a mirror, we quickly realize that the motion of our reflected image is reversed as a function of our muscle movements, and when we observe ourselves on camera (e.g., a security camera in a grocery store), we can quickly pick ourselves out from a crowd simply by looking for the motions that correlate with our actions. We can even understand the effects of our actions under complex optical transformations, such as in the case of a surgeon performing a procedure using a laparoscope. In short, we can quickly discover our own ``end-effector'' (either our own hand, or even a tool) and visually guide it to perform a desired task.

The ability to quickly acquire visual servoing skills of this sort under large viewpoint variation would have substantial implications for autonomous robotic systems: if a robot can learn to quickly adapt to any viewpoint, it can be dropped without any calibration into novel situations and autonomously discover how to control its joints to achieve a desired servoing goal. However, this poses a substantial technical challenge: the problem of discovering how the controllable degrees of freedom affect visual motion can be ambiguous and under-specified from a single frame. Consider the two scenes shown in the lower half of Figure~\ref{fig:teaser}. Which way should the robot command its end-effector to move in order to reach the unseen query object? In the two settings, the action in the robot's (unknown) coordinate frame has almost the opposite effects on the observed image-space motion.
After commanding an action and observing the movement, it is possible to deduce this relationship. However, identifying the effect of actions on image-space motion and successfully performing the servoing task requires a robust and sophisticated perception system augmented with the ability to maintain a memory of past actions.

In this paper, we show that view invant visual servoing skills can be learned by deep neural networks, augmented with recurrent connections for memory. In classical robotics, visual servoing refers to controlling a robot in order to achieve a positional target in image space, typically specified by positions of hand-designed keypoint features~\cite{wilson1996relative,hutchinson1996tutorial}. We instead take an open-world approach to visual servoing: the goal is specified simply by providing the network with a small picture of the desired object, and the network must select the actions that will cause the robot's arm to reach that object, without any manually specified features, and in the presence of severe viewpoint variation. The result is a visual servoing mechanism that is independent of the camera viewpoint and can servo to user-chosen objects. This mechanism automatically and implicitly learns to identify how the actions affect image-space motion, and can generalize to novel viewpoints and objects not seen during training. We train our view-invariant controller primarily in a randomized simulation setup where we generate diverse scenes and provide reinforcement supervision. We then transfer this controller to the real world by adapting the visual features using a small number of labeled real-world images.

The main contribution of our work is a novel recurrent convolutional neural network controller that learns to servo a robot arm to previously unseen objects while it is invariant to the viewpoint. To learn perception and control for such view invariant servoing, we propose a training procedure that uses demonstrated trajectories in simulation as well as a reinforcement learning (RL) based policy evaluation objective. We use a small amount of real-world images to adapt the visual features to enable successful servoing on a real robotic arm while the overwhelming majority of training data is generated in a randomized simulator. Our experimental results evaluate the importance of recurrence for visual servoing on an extensive simulated benchmark and show that incorporating the value prediction function improves the results. We also evaluate the effectiveness of our system in several real-world servoing scenarios both quantitatively and qualitatively.

%% file: related.tex
\begin{figure*}[t]
\begin{center}
\includegraphics[width=.98\linewidth]{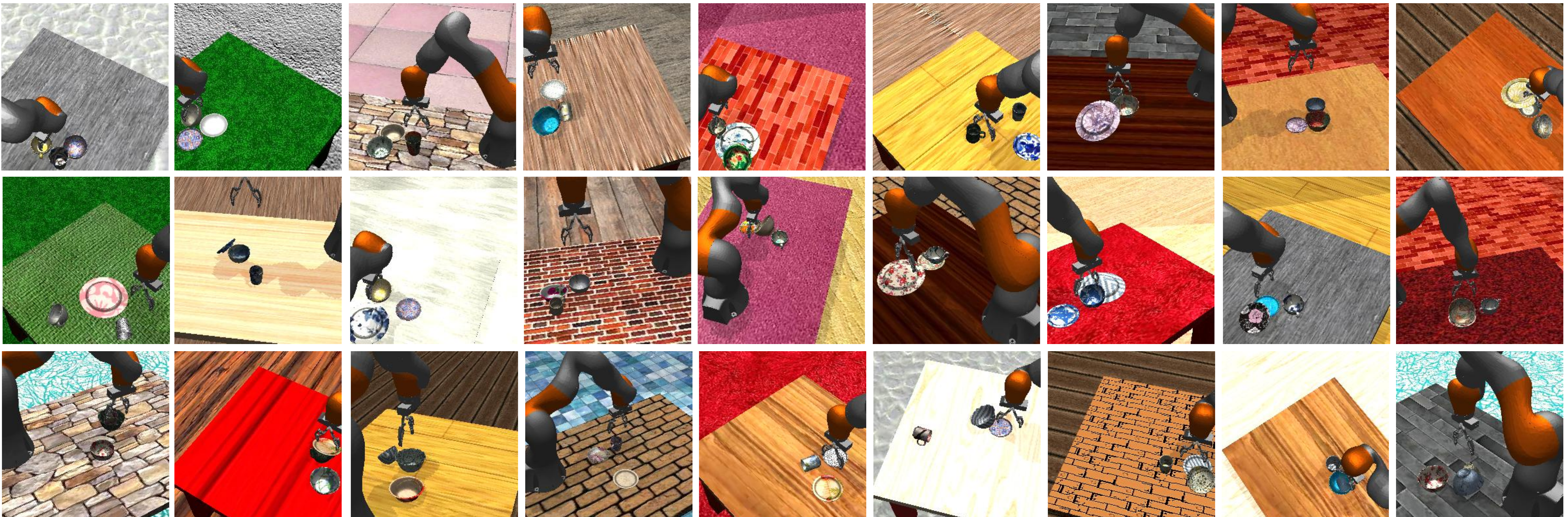}
\end{center}
\vspace{-.15in}
   \caption{We use randomized simulated scenes, as well as randomization of viewpoints, in order to train a recurrent controller in simulation for viewpoint invariant visual servoing.}
\label{fig:simtxtr}
\vspace{-.15in}
\end{figure*}

Visual servoing has a long history in computer vision and robotics~\cite{chaumette2006visual, hutchinson1996tutorial}. Our proposed visual servoing method aims to address a very similar problem, but differs in several important aspects of the visual servoing problem formulation, with the aim of producing a method that is more general and practical for open-world settings.
We depart from a common assumption that the camera intrinsics and extrinsics are calibrated~\cite{mohta2014vision,espiau1992new}, and make no assumptions about the 3D structure of the scene~\cite{espiau1992new, mohta2014vision,collewet2008visual}. Several prior visual servoing methods also address servoing with uncalibrated cameras~\cite{yoshimi1994active, jagersand1997experimental,massoud2007global}, but all of them address an ``eye-in-hand'' setting, where the goal is to servo the camera toward a target view by using previously known geometric features and estimate the image Jacobian within an image based visual servoing setup. In contrast, our visual servoing setting involves servoing a robotic arm to a visually indicated target, provided via a query image while the camera viewpoint is unknown and can change between trials. Crucially, this query image is \emph{not} the desired image that the camera should see (as in the eye-in-hand setup), but rather an object that arm should approach, while the camera observes the scene from an unknown  viewpoint. This requires the servoing mechanism to learn to match visual features between the query object and current observation, recognize the motion of the arm, and account for differences in viewpoint between trials.

Specifying the target by providing an image of the query object, instead of specifying low-level keypoints, is most most similar to photometric visual servoing~\cite{caron2013photometric}. However, while photometric visual servoing aims to match a target image (e.g., by moving the camera), our method aims to direct the arm to approach the visually indicated object. The query image provides no information about \emph{how} to approach the object, just which object is desired. Our model must therefore both localize the object and direct the robot's motion.

Similarly to recent work on self-supervised robotic learning~\cite{levine2016end,levine2016learning,agrawal2016learning,gandhi2017learning,pinto2016supersizing}, our method uses observed images and self-supervision to train deep convolutional networks to predict action-conditioned task outcome. However, in contrast to these prior methods, our camera viewpoint is not fixed and can change drastically from one episode to another. Our approach incorporates fast adaptation via recurrence to adapt the visual servo to a novel viewpoint within a single episode of task execution, in addition to an outer-level self-supervised learning process performed with conventional gradient descent.

The use of recurrent networks for control has previously been explored in a number of works on reinforcement learning, including methods for navigation~\cite{mnih2016asynchronous,oh2016control},
continuous control~\cite{heess2015memory,zhang2016learning}, and simulation to real world transfer~\cite{peng2017sim}. However, to our knowledge, no prior method has demonstrated that recurrence can be used to enable real-world robotic visual servoing from novel viewpoints. The closest work to this topic has taken a system identification approach for unknown physical parameters, such as masses and friction parameters~\cite{yu2017preparing} and does not use either or image observation or recurrence. Also~\cite{peng2017sim} does not incorporate image observations.

We use randomized simulated experience to train our visual servoing system. In order to use it for visual servoing in the real world, we also introduce an adaptation procedure based on finetuning of the visual features with an auxiliary objective. Most prior approaches to domain adaptation either train domain invariant representation~\cite{long2015learning, bousmalis2016domain, ganin2016domain,bousmalis2017using}, learn a representation transformation from one domain to another via transfer learning~\cite{gopalan2011domain,tzeng2015adapting,rusu2016sim}, or employ domain randomization in simulation~\cite{sadeghi2017cadrl,tobin2017domain} which produces robust models that can generalize broadly and can directly be deployed in the real world. Our approach combines domain randomization with transfer learning: we learn the controller entirely in a randomized simulation environment and then finetune only the visual features with a small amount of real world images, effectively transforming the model's representation into the real-world domain. We show that our final finetuning procedure produces an effective visual servoing mechanism in the real world, even though the recurrent motor control layers are not finetuned on the real-world data.

%% file: approach.tex
\label{sec:rcm}

Our aim in this paper is to study visual servoing scenarios that require generalization to new viewpoints, in order to understand whether deep recurrent neural networks can learn to implicitly ``self-calibrate'' and understand the relationship between motor commands and motion in the image. To this end, we set up a visual servoing scenario where a robot arm must reach for objects in the world, using monocular camera observations from an arbitrary viewpoint. The goal is indicated by an image of the query object, and the network must both figure out where this object is in the image and how to actuate the robot arm in order to reach it.

The principal challenge in this problem setup comes from the inherent ambiguity over the motion of the arm in the image in terms of the actions. Most standard visual servoing methods assume knowledge of the Jacobian -- the relationship between actions and motion of desired visual features. Our approach not only has no initial knowledge of the Jacobian, but it does not even have any prior visual features, and must learn everything from data. Determining the right actions from a single image is generally not possible. Instead, we must incorporate temporal context, using the outcomes of past actions to inform future ones. To that end, we use a recurrent neural net (RNN), whose internal state is trained to extract and capture knowledge about hand-eye coordination during the course of a single episode. Our network must take a few initial actions, observe their outcomes, and implicitly ``self-calibrate'' to understand how actions influence image-space motion.  We train this recurrent controller over a large number of randomized scenes and show that it generalizes to novel scenes, with variability in both viewpoint and object appearance.

Another challenging aspect of our problem is the visual complexity of the scene. Our setup includes varied scene layouts with a wide variety of objects and we render the table, plane and object with random textures as well as random lighting conditions in a 3D simulator. For each episode, we randomly select the query object and distractor objects 3D shapes and place them on a random location on the table and with random orientation. This setup enforces the model to learn to distinguish between objects as it has to reach for a query object. As such it needs to implicitly perform object localization in 3D. 

In this section, we describe the design of our visual servoing model and recurrent controller, which must observe both the current image and a closely cropped image of the query object, and produce the action as output. In Section~\ref{sec:training}, we will describe how the model is trained using a combination of action supervision gathered from demonstrated trajectories in simulation and reinforcement learning based value function prediction. In Section~\ref{sec:sim_real}, we will describe how we train in simulation and transfer to a real world robot. Fig.~\ref{fig:architecture} illustrates our full network architecture.

We denote our controller model as $\pi_\theta$. This model is a function, with parameters $\theta$, that accepts as input the current image observation $o_t$ and the query image $q$, as well as the previous internal state $h_{t-1}$ representing the memory. The previously chosen action $a_{t-1}$ is also provided, so that it can infer how actions affect image-space motion. The output consists of the action $a_t$ and the new internal state $h_t$, such that the policy is defined as
\begin{equation}\label{eq:model}
a_{t+1}, h_{t+1} = \pi_{\theta}(o_t, a_{t-1}, q, h_t) 
\end{equation}
We implemented recurrence in our controller using an LSTM, and the action is defined as a displacement $a=(\partial x, \partial y, \partial z)$ of the end effector of the arm in the robot's frame of reference (which is not known to the model). For the purpose of exploration, the policy that is used to collect experience during training is stochastic, and corresponds to a Gaussian with mean given by the model output. When the model is used to select actions for $T$ steps, it produces a sequence of observation and actions $\tau=(o_1, a_2, \cdots, o_T, a_T)$. 

The observation $o_t$ and query image $q$ are processed with separate convolutional stacks based on the VGG16 architecture~\cite{simonyan2014very}, with $o_t$ having an input size of $256 \times 256$ and $q$ resized to $32 \times 32$. These networks are trained from scratch, without pretraining. This produces vector representations of both images, that are concatenated with each other and the previous action $a_{t-1}$, which is transformed via a fully connected ReLU layer into a 64-dimensional feature vector. The observation embedding at step $t$, the query embedding and the action embeddings are concatenated in one vector as input to the recurrent motor control system, which uses a single-layer LSTM with $512$ units~\cite{hochreiter1997long}. The state of this LSTM corresponds to the memory $h_{t-1}$, which allows the model to capture information from past observations that is needed to perform implicit calibration.

\begin{figure}
\begin{center}
\includegraphics[width=.98\linewidth]{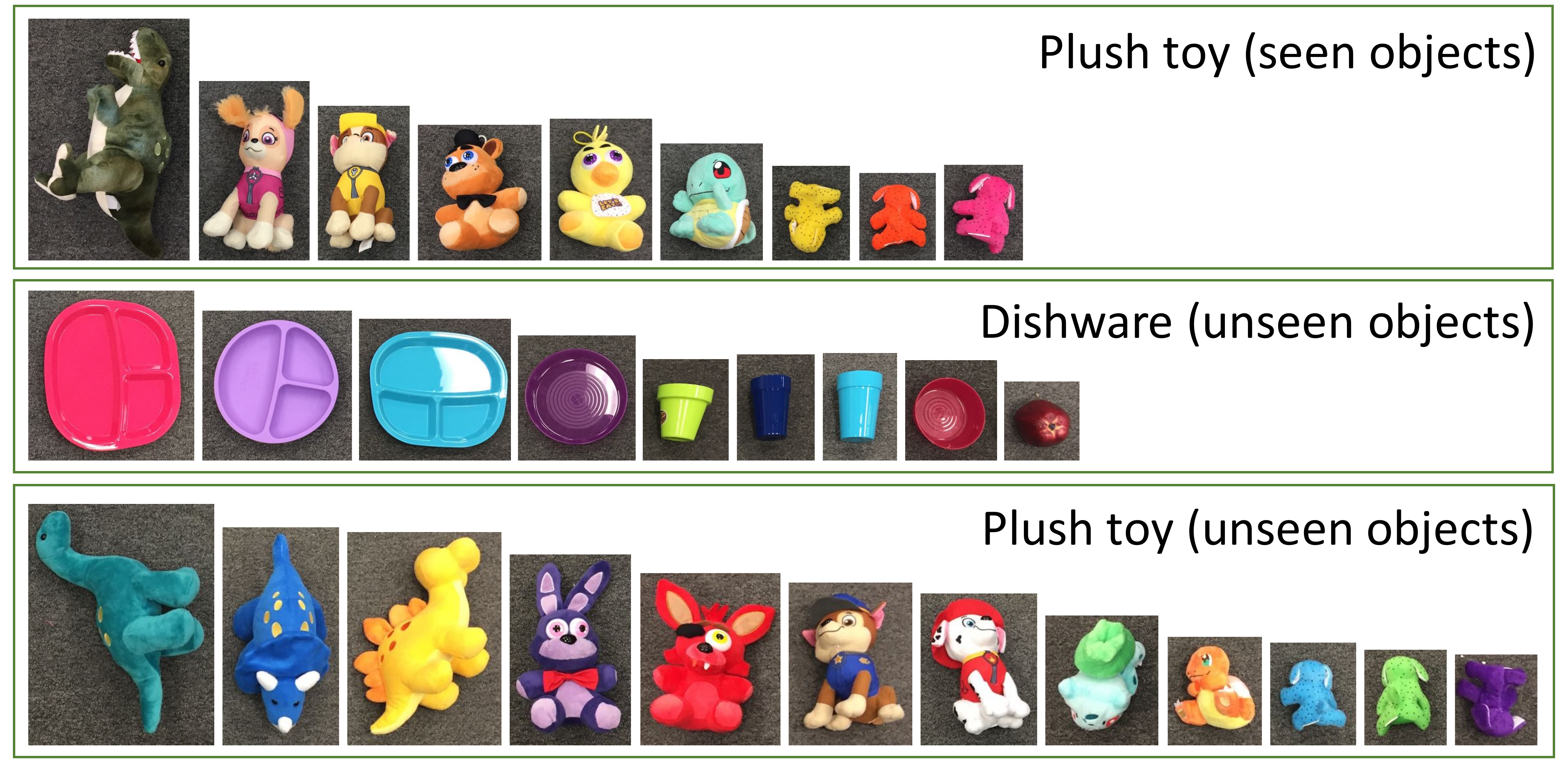}
\end{center}
\vspace{-.15in}
   \caption{The set of seen and unseen object used in the real-world experiments. The seen plush toys are used for adapting the visual layers to natural images, while the unseen objects are used for testing.}
\label{fig:test_objects}
\vspace{-.15in}
\end{figure}

\section{Training}
\label{sec:training}

Our visual servoing model is trained with a combination of supervised learning, which is analogous to learning from demonstration, and outcome prediction, which corresponds to a type of reinforcement learning. In our implementation, the model is trained entirely in simulation (see Sec.~\ref{sec:sim_real}), which provides full access to object locations and robot states. This allows us to produce supervision that corresponds to ground truth actions or synthetic ``demonstrations.''\footnote{The model could also be trained in the real world, with ground truth actions provided by actual human or teleoperated demonstrations, but we opted for simulated training in order easily collect huge amount of highly varied data.} These demonstrations directly supervise the action output $a_t$. However, this supervision does not directly teach the network about the long term effects of an action, and we found it beneficial to also augment the training process with a value prediction loss, which corresponds to reinforcement learning. This loss trains the model to also predict the state-action value function associated with each action using multi-step Monte Carlo policy evaluation, which is the reward that the model expects to obtain for the entire episode if it takes a particular action in a particular state and then follows its own policy. As shown in our experiments, this RL loss leads to improved performance, since the resulting internal representations become better adapted to the long-term goal of the task.

\subsection{Learning from Synthetic Demonstrated Trajectories}

We synthesize the strongly supervised training data by generating a large set of episodes with varied camera locations, objects, and textures. Details of the simulation setup are provided in Sec.~\ref{sec:sim_real}. Each episode contains ground truth actions that servo the arm to the query object, perturbed by injecting Gaussian noise to provide some degree of exploration, which we found beneficial for producing robust polices. The training loss corresponds to the sum of squared Euclidean distances between the output action and the vector from the end-effector to the target object. Using $x_t$ to denote the end-effector position and $y$ to denote the query object position, the loss can be written as
\begin{equation}\label{eq:loss}
    Loss = \sum_{t=1}^{T}||(y - x_{t}) - a_{t}||^2.
\end{equation}
To keep the action magnitudes within a bound, we learn normalized action direction vectors and use constant velocity. Note that the target is the same at each step, and corresponds to the true object location. The sampled trajectories provide starting points of the arm and random past actions from which the model needs to recover. After unrolling the policy and formulating the above loss we use stochastic gradient descend over the parameters $\theta$ to minimize $Loss$. After our model converges, we generate additional on-policy samples: we use the current policy to generate on-policy trajectories, and then label them with the ground truth actions, following the DAgger framework~\cite{rgb-rilsp-11}. This new data is then used to train the model to convergence. We repeat this procedure for two iterations.

\subsection{Learning the Value Function}

The supervised learning procedure provides detailed supervision, which means that it can quickly lead to a reasonable policy, but it is also myopic, in the sense that it does not consider the long term effects of an action. We found that the final performance of our model could be improved by also incorporating a reinforcement learning objective, in the form of state-action value function prediction, which is also known as the Q-function~\cite{sutton1998reinforcement}. This allows us to then select the action that minimizes the predicted long term reward. We formulate a reward function that indicates whether the arm has reached the target, such that $r(s_t,a_t) = 1$ if the arm is at the target at step $t$, and $r(s_t,a_t) = 0$ otherwise. Here, we use $s_t$ to denote the (unobserved) underlying state of the system, which includes the arm pose and query position. The target Q-values are then computed according to
\[
Q(s_t,a_t) = r(s_t,a_t) + E_{\tau \sim \pi_\theta}\left[ \sum_{t'={t+1}}^T \gamma^{t'-t} r(s_{t'},a_{t'}) \right]
\]
where $\gamma$ is a discount factor. These target values are used with a squared error regression loss applied to a second head of the model (see Fig.~\ref{fig:architecture}). The rewards and target Q-values are computed along trajectories sampled by running the policy. In practice, we found it beneficial to unroll the policy multiple times from each visited state and action and average together the corresponding returns to estimate the expectation with respect to $\pi_\theta$. This corresponds to multi-step Monte Carlo policy evaluation~\cite{sutton1998reinforcement}, which provides a simple and stable method for learning Q-functions~\cite{sadeghi2017cadrl}.
We can choose actions according to this Q-function by optimizing with respect to the input action $a_t$. In our implementation, we use the cross-entropy method (CEM)~\cite{rubinstein1999cross} to perform this optimization, which provides a simple gradient-free optimization procedure for choosing actions that maximize predicted Q-values.

\section{Simulated Training and Transfer}
\label{sec:sim_real}
One of the main challenges in robot learning is data collection. While deep models are shown to work well on huge amount of data, collecting robot data is time consuming and expensive which results in learning challenges and models with low generalization. Therefore, we train our controller in simulation, which allows us to generate a large, diverse range of scenes with different viewpoints, objects and color, and easily obtain the supervision needed to train our model efficiently. We use domain randomization to learn a robust visual model. However, too boost the performance in the real world, we use visual adaptation with small amount of real world images which we also describe in this section.

\subsection{Simulated Environment}\label{sec:sim_env_train}
We use the Bullet physics engine for simulation~\cite{coumans2017}, with a simulated 7 DoF Kuka IIWA arm and a variety of objects placed on a tabletop surface in front of the arm. The objects are randomly selected from a set of 50 scanned objects representing various dishware -- plates, mugs, cups, bowls, glasses, and teapots. The objects are dropped on the table, so that their pose is randomized. We also randomize textures, lighting, and the appearance of the table and ground plane. This randomization procedure serves two important purposes: first, it forces the controller to learn the underlying geometric patterns that are actually important to the task, rather than picking up on extraneous aspects of object appearance that might correlate with actions, and second, it serves to enable the model to generalize more easily to real-world scenes, by essentially forcing it to solve a harder generalization task (widely different appearances), as discussed in prior work~\cite{sadeghi2017cadrl}. Each simulated trial consists of a random camera viewpoint, up to three randomly selected objects in random poses, and randomized appearance parameters. Figure~\ref{fig:simtxtr} shows examples of our randomized simulation environment from various robot camera viewpoints.

\begin{figure*}[t]
\begin{center}
\includegraphics[width=.98\linewidth]{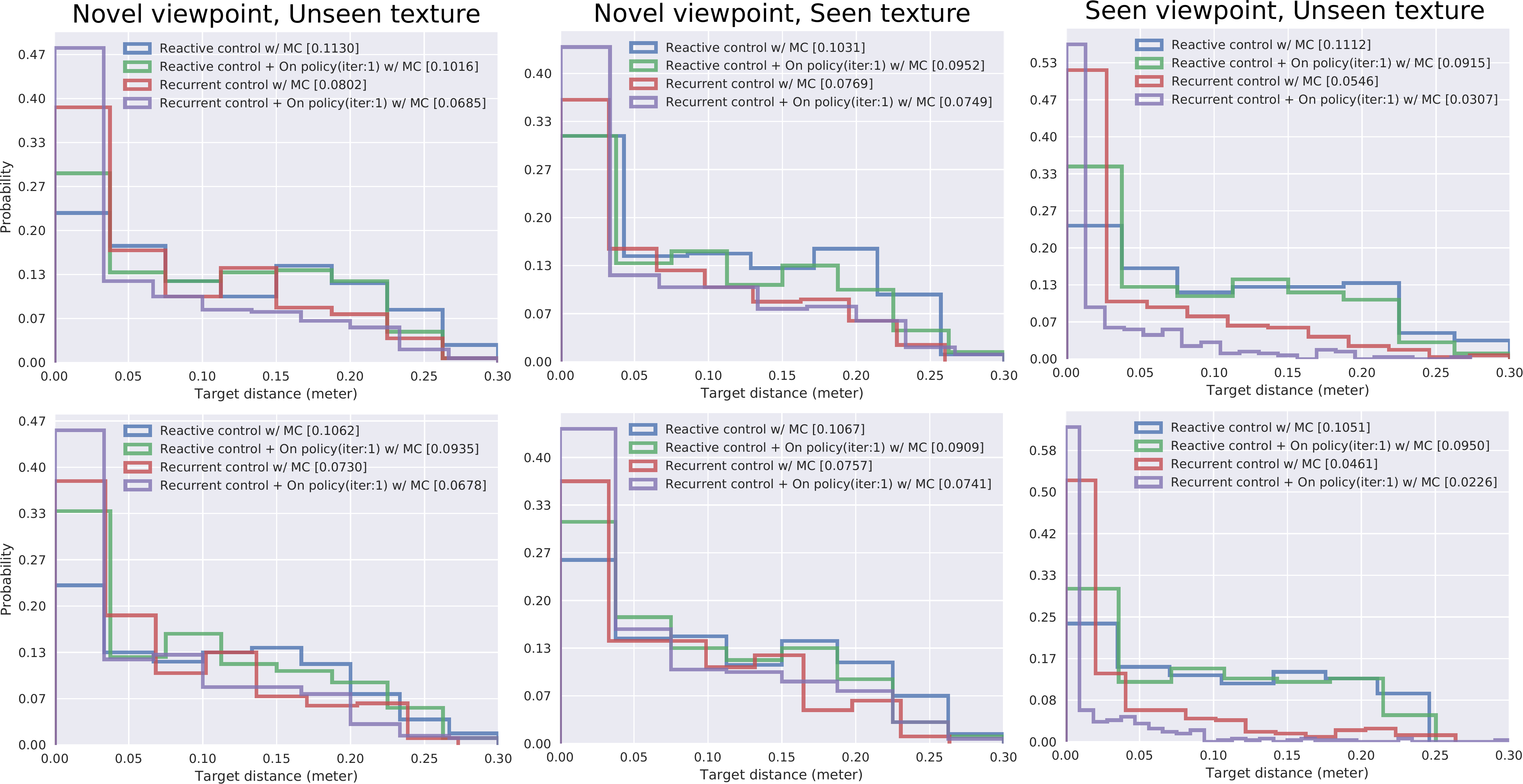}
\end{center}
\vspace{-.15in}
   \caption{Comparing recurrent control vs reactive control in test scenarios with different levels of difficulty. Top row: test scenarios with three random objects. Bottom row: test scenarios with two random objects}
\label{fig:recurrentvsff}
\vspace{-.15in}
\end{figure*}

\subsection{Adaptation to the Real World}
In order to enable our controller to perform visual servoing in the real world, with a real robotic arm, we must adapt the parameters of the model to handle real images. The randomization procedure described in the previous section already provides some degree of generalization. In fact, prior methods observed that simply randomizing the environment can enable manipulation of simple geometric shapes in the real world~\cite{tobin2017domain}. However, in order to enable the model to interact with realistic natural objects, such as those shown in Figure~\ref{fig:test_objects}, we found that an additional adaptation step was required. Obtaining either ground truth actions or rewards in the real world would require costly manual labeling. Instead, we can leverage the fact that the motor control portion of our model can remain largely unchanged between simulation and the real world, and only the visual features must be adapted. To that end, we can use a weaker form of supervision to finetune only the convolutional layers of the model.

To that end, we use the auxiliary adaptation loss at the last layer of the visual stack (see Fig.~\ref{fig:architecture}, top-left), which simply predicts the presence or absence of the query object on a $8 \times 8$ grid overlaid on the image, by using convolutionally computed logits at the last convolutional layer. These logits are fed into a cross entropy loss which is used to finetune the vision stack. As training data for this loss, we use 22 sequences of the arm executing random actions, and we annotate the first frame in each video with bounding boxes for the objects that are present on the table. This resulted in a total of 76 bounding boxes. Some of these scenes are shown in Figure~\ref{fig:realdata}. Since the episodes remain stationary during each episode, we can propagate the labels automatically through each sequence. The actual loss is constructed by sampling a batch of sequences, and for each sequence sampling one object to use as the query object by cropping out one bounding box. To make our localization robust against object poses, for each sequence we randomly select the query image input from a pool of query images of the same query object category. The loss then describes the error in localizing that query object in the spatial image frame.

\subsection{Implementation Details}

We implement our model in TensorFlow~\cite{abadi2016tensorflow}. We use a buffer of one million unrolls for each policy optimization iteration. We use the Adam optimizer~\cite{kingma2014adam} for training, with a learning rate of $1.5e-5$ and an exponential decay schedule. The training in each round converges after one million steps.

%% file: exp.tex
Our experimental evaluation consists of a detailed simulated comparison with alternative approaches, as well as a real-world evaluation with a robotic arm to study generalization to real-world situations. Standard visual servoing methods in the literature are not directly applicable to our problem setting, since we do not assume knowledge about the camera position or the action to image Jacobian, and the goal is specified simply by providing a picture of a desired object that the arm should reach for. Instead, we compare to ablated variants of our method. To evaluate the importance of memory for learned, implicit self-calibration, we compare to a non-recurrent, feedforward visual servo, trained in exactly the same way as our method. To evaluate the importance of combining supervised learning from demonstrated trajectories with RL based value prediction, we also compare to a recurrent model that is not trained to predict value. 

\subsection{Simulated Reaching}

In our simulation experiments, we aim to answer the following questions: (1) How effective is our proposed recurrent controller compared to a feedforward reactive policy? (2) How well does the model deal with viewpoint changes and visual variation? (3) What is the benefit of incorporating on-policy data? (4) How beneficial is the value prediction objective?

\noindent{\bf Model setup:} In addition to the recurrent controller, we also train a reactive feed-forward policy. For the reactive policy we use a feed-forward network which has same layers as in Fig.~\ref{fig:architecture}, but has two fully connected layers instead of the recurrent LSTM layer.

\begin{figure*}[t]
\begin{center}
\includegraphics[width=.98\linewidth]{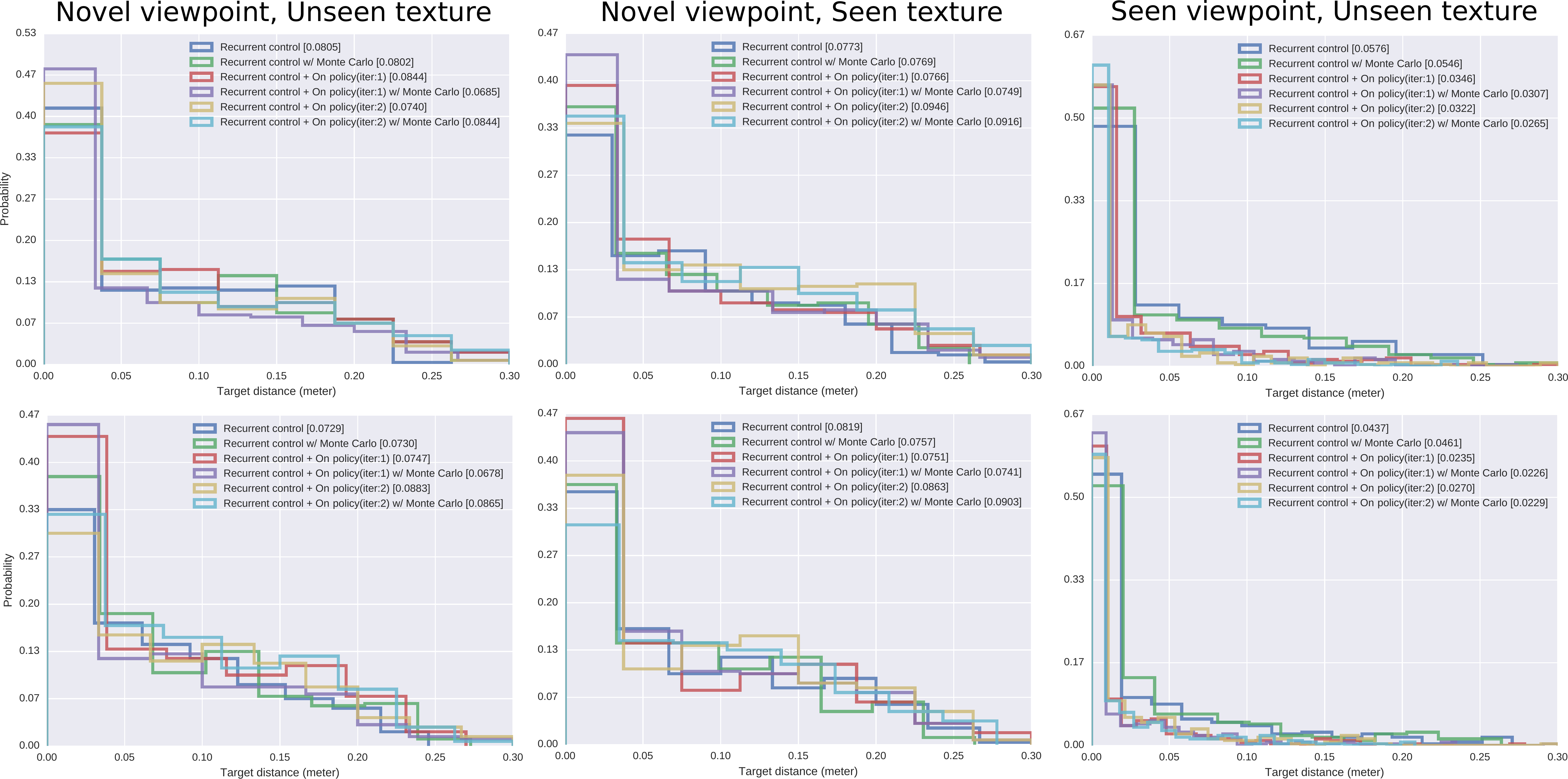}
\end{center}
\vspace{-.2in}
\caption{Comparison for different numbers of iterations of on-policy data collection, as well as the benefit of the value prediction objective by using Monte-Carlo policy evaluation. Top row: test scenarios with three random objects. Bottom row: test scenarios with two random objects. Left, middle and right show test scenarios with different levels of difficulty.}
   \vspace{-.05in}
\label{fig:mcnomc}
\end{figure*}

\noindent{\bf Test setup:}
For testing, we generate new simulated scenarios using the same randomization procedure as described in Section~\ref{sec:sim_env_train}. We use objects that were not seen during training, with two or three objects in each scene. The test viewpoints are sampled uniformly at random in a region around the workspace, with the orientation chosen to always point toward the table, such that the query object is visible. We randomly select of the viewpoints for training episodes, while keeping a held-out set of test viewpoints to test generalization. We aim to evaluate the performance of our method on test scenarios with different levels of difficulty. To do this, we consider three different types of test scenes: (a) Novel textures and novel viewpoints. (b) Previously seen textures and novel viewpoints. (c) Novel textures and previously seen viewpoints.

\noindent{\bf Evaluation criteria:} In each simulation experiment, we run the policy for 300 trials, each with a fixed length of 10 steps. At the end of each trial, we compute the Euclidean distance of the robot's end-effector to the query object, using the closest points on the arm and the object mesh. This metric is in meters, and can be zero when the arm touches the object. We report the average value of the distance to the query object over the last time steps of the 300 test trials.
We visualize the distribution of the final distances attained at the end of the trials in Figure~\ref{fig:recurrentvsff}, and report the average distances in Table~\ref{tab:sim_exp_avg}.

\begin{table*}
\begin{small}
\begin{center}
\caption{\small Average distance to target in meter for various test settings with two and three objects scenes. (VP: Viewpoint, T: Texture) \vspace{-.01in} \label{tab:sim_exp_avg}}
\renewcommand*{\arraystretch}{0.9}
\begin{tabular}[10pt]{lcccccc}
\toprule
  \multirow{3}{*}{}& \multicolumn{3}{c}{Three objects} & \multicolumn{3}{c}{Two objects} \\\cline{2-7}
 
 &{Novel VP} & {Novel VP} & {Seen VP} &{Novel VP} & {Novel VP} & {Seen VP} \\
  & {Unseen T} & {Seen T } & {Unseen T } & {Unseen T} & {Seen T } & {Unseen T } \\
\midrule
{Reactive w/ MC}             & {0.1130} & {0.1031} & {0.1112} & {0.1062} & {0.1067} & {0.1051}  \\
{Reactive + On policy w/ MC} & {0.1016} & {0.0952} & {0.0915} & {0.0935} & {0.0909} & {0.0950}  \\
{Recurrent w/ MC}            & {0.0802} & {0.0769} & {0.0546} & {0.0730} & {0.0757} & {0.0461}  \\
{Recurrent + On policy w/ MC}  & {0.0685} & {0.0749} & {0.0307} & {0.0678}  & {0.0741} & {0.0226} \\
\bottomrule
\vspace{-.25in}
\end{tabular}
\end{center}
\end{small}
\end{table*}

\begin{table*}
\begin{small}
\begin{center}
\caption{\small Average distance to target in meter for evaluating the effect of the value prediction loss by using Monte-Carlo policy evaluation (MC) and on-policy data. (VP: Viewpoint, T: Texture) \vspace{-.01in} \label{tab:sim_MC_onP_ablation}}
\renewcommand*{\arraystretch}{0.9}
\begin{tabular}[10pt]{lcccccc}
\toprule
  \multirow{3}{*}{}& \multicolumn{3}{c}{Three objects} & \multicolumn{3}{c}{Two objects} \\\cline{2-7}
 
 &{Novel VP} & {Novel VP} & {Seen VP} &{Novel VP} & {Novel VP} & {Seen VP} \\
  & {Unseen T} & {Seen T } & {Unseen T } & {Unseen T} & {Seen T } & {Unseen T } \\
\midrule
{Recurrent}                                       & {0.0805} & {0.0773}& {0.0576} & {0.0729} & {0.0819} & {0.0437} \\
{Recurrent w/ MC}                             & {0.0802} & {0.0769}& {0.0546} & {0.0730} & {0.0757} & {0.0461} \\
{Recurrent + On policy}                      & {0.0844} & {0.0766}& {0.0346} & {0.0747} & {0.0751} & {0.0235} \\
{Recurrent + On policy w/ MC}            & {0.0685} & {0.0749}& {0.0307} & {0.0678} & {0.0741} & {0.0226} \\
{Recurrent + On policy(iter:2)}            & {0.0740} & {0.0946}& {0.0322} & {0.0883} & {0.0863} & {0.0270} \\
{Recurrent + On policy(iter:2) w/ MC}  & {0.0844} & {0.0916}& {0.0265} & {0.0865} & {0.0903} & {0.0229} \\

\bottomrule
\vspace{-.2in}
\end{tabular}
\end{center}
\end{small}
\end{table*}

\noindent{\bf Reactive vs. recurrent policies:} As illustrated by the final distributions, the reactive policy is substantially less proficient at reaching the query objects. When the testing scenario is the most challenging, with novel textures and novel viewpoints, and without any on-policy data collection, the average final distance obtained by the reactive policy is 0.11m, while the recurrent policy reaches an average final average distance of 0.08m. Incorporating on-policy data for training our proposed approach results in a final distance of 0.07m in the novel viewpoint and unseen texture condition. The results also indicate that the novel camera viewpoints are indeed more challenging when it comes to generalization.

\noindent{\bf The effect of using on-policy data:}
The effect of using different numbers of iterations of on-policy training is shown in Figure~\ref{fig:mcnomc} and also summarized in Table~\ref{tab:sim_MC_onP_ablation}. Performing two iterations of retraining with on-policy data produces the best performance on the seen viewpoints and unseen texture scenarios, while resulting in poorer performance in the scenarios with novel viewpoints. On the other hand, using one iteration of retraining with on-policy data improves the performance in all scenarios. This result suggests that one iteration of on-policy data collection can help to address the distribution mismatch problem, though additional iterations can potentially result becoming more specialized to the training viewpoints. Therefore, if the task emphasis is on mastery, using more on policy data can improve performance.

\begin{figure}[t]
\begin{center}
\includegraphics[width=.8\linewidth]{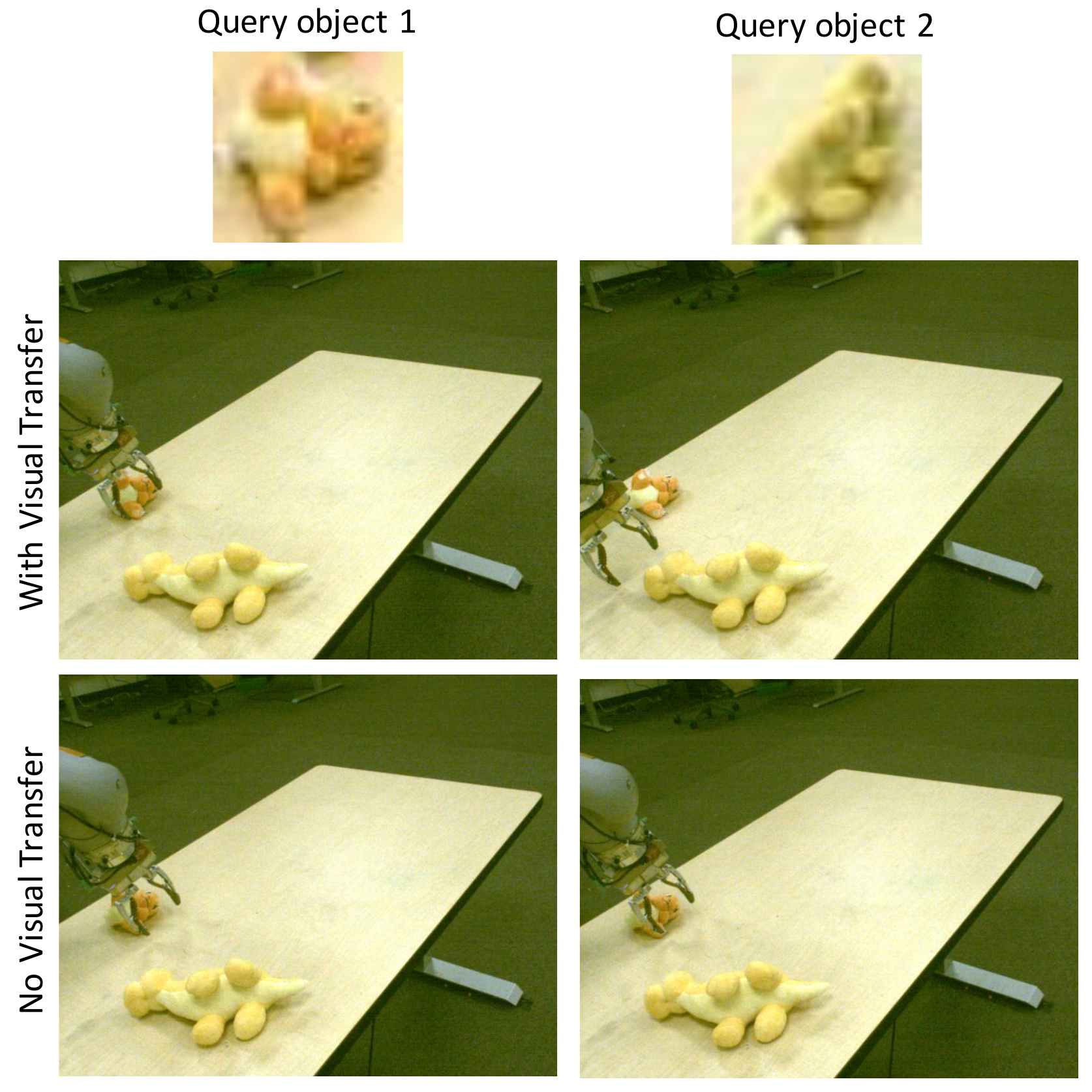}
\end{center}
\vspace{-.15in}
   \caption{The network trained with only simulated data becomes confused between two objects with similar color and fails in the reaching task, while the visually adapted network can distinguish between the two object.}
\label{fig:realqual}
\vspace{-.2in}
\end{figure}

\noindent{\bf The effect using Monte Carlo policy evaluation for value prediction :}
To evaluate the effectiveness of the value prediction loss which uses multi-step Monte-Carlo (MC) policy evaluation, we conducted simulated experiments with and without the value prediction loss. When the value prediction loss is used, it is  denoted by w/Monte Carlo in Figure~\ref{fig:mcnomc}. In these experiments, we compute the best action at each time step by using the action prediction output of the model, and then use CEM to perturb this action with Gaussian noise with standard deviation $\sigma=0.003$ to generate $150$ candidate actions. We then use the value prediction head to evaluate each of these candidate actions and sort them based on their value. The executed action is sampled at random from the top 5 highest scoring actions.
The results in Figure~\ref{fig:mcnomc} shows that, in most conditions, incorporating the value prediction head which is trained using Monte-Carlo return estimates results in improved performance.

\begin{figure}[t]
\begin{center}
\includegraphics[width=.98\linewidth]{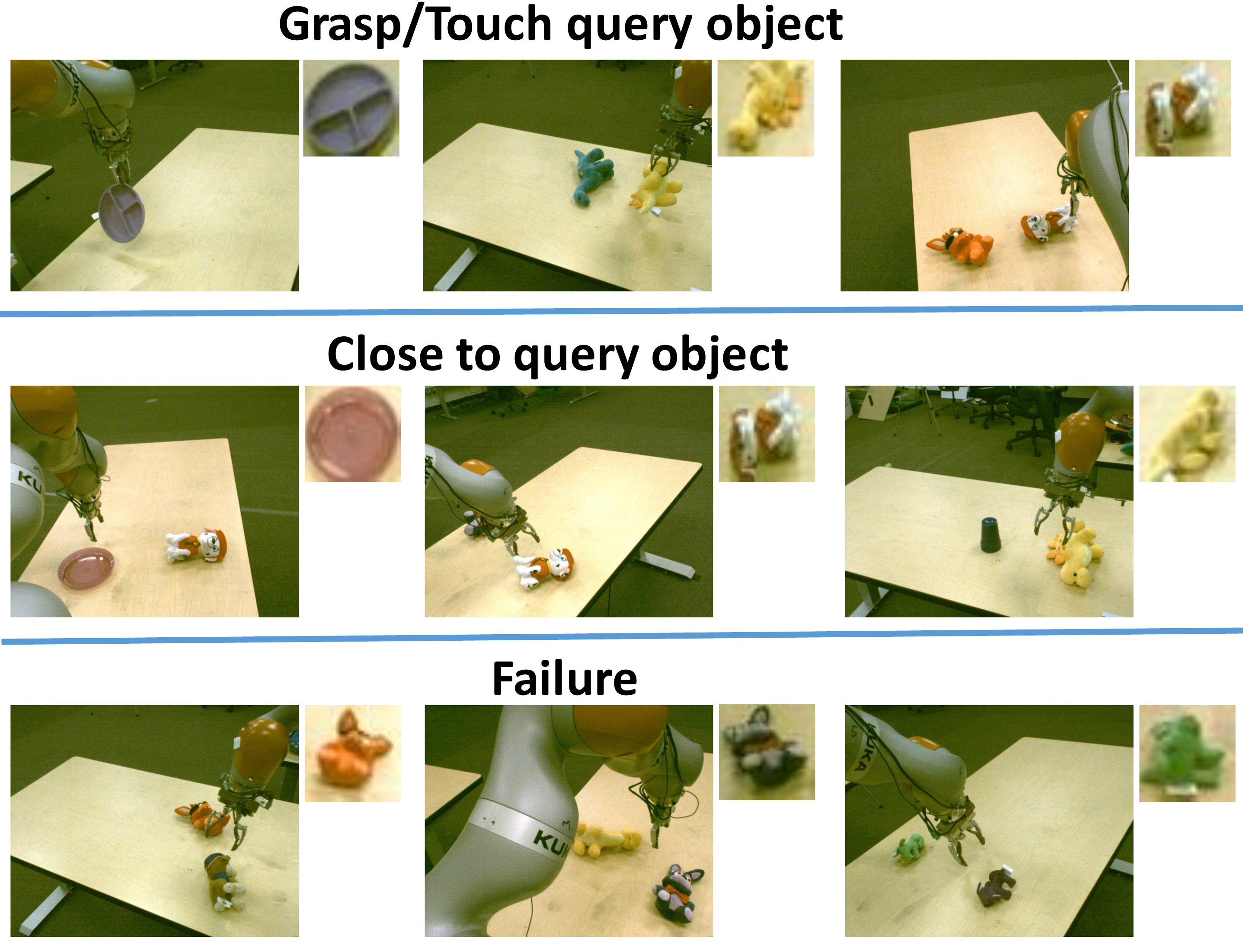}
\end{center}
\vspace{-.2in}
   \caption{A successful reach occurs if the gripper touches the query object or gets very close to it. Failure examples include cases where the controller is confused between multiple objects, and ends up with a far distance from the query object or approaching the wrong object.}
   \vspace{-.15in}
\label{fig:real_analysis}
\end{figure}

\begin{figure*}[t]
\begin{center}
\includegraphics[width=.9\linewidth]{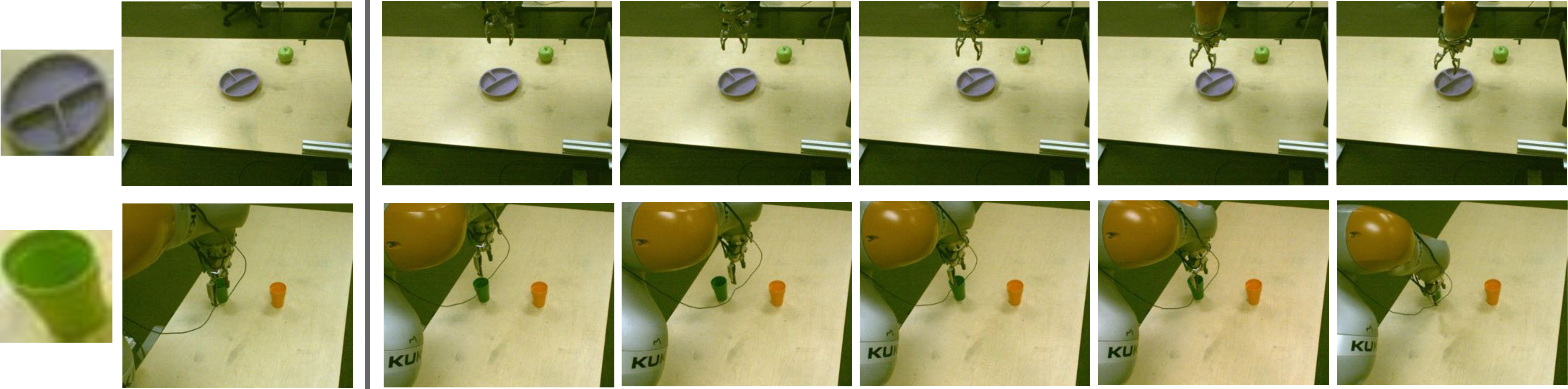}
\end{center}
\vspace{-.15in}
   \caption{In both scenarios, the arm successfully reaches the object. Note that, in the second sequence, the arm first moves to the right, and then observes the effect of this action and corrects, moving toward the query object. This suggests that the controller can observe action outcomes and incorporate these observations to correct servoing mistakes.}
\label{fig:real_self_calibration}
\vspace{-.1in}
\end{figure*}

\begin{table*}
\begin{small}
\begin{center}
\caption{\small Real world reaching task results with novel viewpoints (Percentage of successful trials).\vspace{-.05in} \label{tab:real_exp}}
\renewcommand*{\arraystretch}{0.9}
\begin{tabular}[10pt]{lcccccc}
\toprule
  \multirow{3}{*}{}& \multicolumn{3}{c}{Simulation only controller} & \multicolumn{3}{c}{Visually adapted controller} \\\cline{2-7}
 
  & \multirow{2}{*}{Success rate} &{Grasp/Touch} & {Close to} & \multirow{2}{*}{Success rate} &{Grasp/Touch} & {Close to query object} \\
    & {} &{query object} & {query object} & {} &{query object} & {query object} \\
\midrule
{One object}        & {88.9} & {55.6}& {33.3} & {94.4} & {61.1} & {33.3} \\
{Two objects}      & {54.1} & {33.3}& {20.8} & {70.8} & {25.0} & {45.8} \\
\bottomrule
\vspace{-.3in}
\end{tabular}
\end{center}
\end{small}
\end{table*}

\subsection{Real-World Robotic Reaching}
We evaluated the generalization capability of our viewpoint invariant visual servoing model on a real-world Kuka IIWA robotic arm. We used two sets of novel objects for the test experiments. The test objects include plush toys with different colors and shapes, as well as different dishware objects, such as cups, plates, and bowls. These objects are shown in Figure~\ref{fig:test_objects}.

\noindent{\bf Quantitative Results:} In our real-world evaluation, we used different objects placed at arbitrary locations on a table, and placed the camera at various locations. Our experiments compare a network with adapted visual layers to one that was trained entirely in simulation without any additional adaptation. The two methods were compared head-to-head on each viewpoint and object arrangement, to provide a low-variance comparison. The tests were divided into scenarios with either one or two objects on the table. Table~\ref{tab:real_exp} summarizes the performance on the real-world reaching task. We performed a total of 42 trials, 18 with single objects, and 24 with two objects. These trials were recorded from a total of 18 camera viewpoints.

To evaluate the performance, we count the number of times the arm moves towards the right query object and reaches it. A successful reach occurs if the gripper touches the object or gets very close to it. If the arm moves in the wrong direction or is confused between the two objects, we count the trial as a failure. We added a fixed procedure at the end of each robot trial to model a pointing action. In this procedure, the gripper is first closed and the arm is pulled up. Then the arm is moved downward and the gripper is opened. Note that, while our model is trained for reaching, and not particularly for the grasping task, using the aforementioned procedure can sometimes result in a grasp if the gripper successfully reaches the query object.   

Figure~\ref{fig:realqual} illustrates examples of successful and unsuccessful reaching attempts. The first row of this figure shows trials where the gripper touches the object or grasps it. The second row shows successful reach attempts where the gripper gets very closed to the query object, and is far from the distractor object. The last row shows several failure cases, where the controller is confused between two objects and the trial ends with the gripper a considerable distance from the query object.

The single object scenarios provide a simpler test setting, where success is mainly dependent on the ability of the method to determine which actions move the arm toward the object, rather than testing the network's ability to localize the right object in the scene. On the other hand, the two-object scenario requires the model to both generalize to a novel viewpoint and distinguish the query object from the distractor. This is significantly more challenging, especially since the test objects differ significantly from the simulated objects seen during training. As seen in Table~\ref{tab:real_exp}, adapting the visual features with a small amount of real-world data substantially improves the performance of the network in both scenarios, with a success rate of 70.83\% in the harder two-object setting. Table~\ref{tab:real_exp} summarizes the outcome of successful trials in more detail and outlines the percentage of these trials that result in the gripper touching or grasping the object.

\begin{figure}[t]
\begin{center}
\includegraphics[width=.9\linewidth]{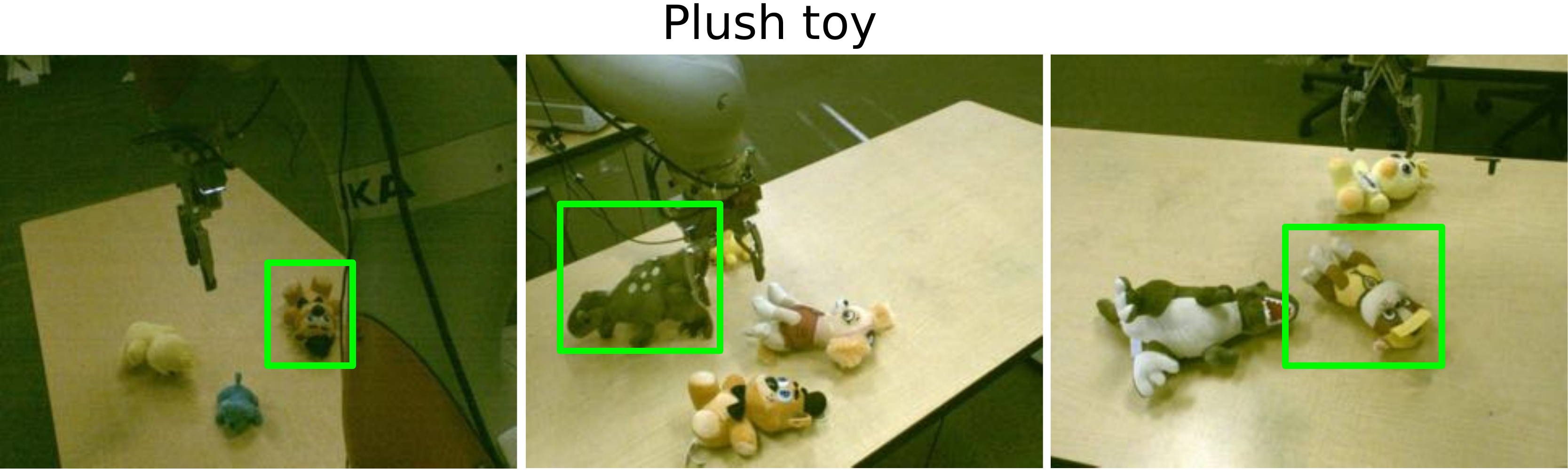}
\end{center}
\vspace{-.15in}
\caption{Examples of real-world scenes used for testing.}
\label{fig:realdata}
\vspace{-.2in}
\end{figure}

\noindent{\bf Qualitative Results: } We visualize two interesting reaching sequences. In Figure~\ref{fig:real_self_calibration} we see successful reaches with exploratory motions, where the arm first moves in the wrong direction, then observes the image-space motion and corrects. In Figure~\ref{fig:realqual}, we observe that the network that is entirely trained in simulation makes more mistakes when the query object and distractor object are visually similar. The network after adaptation is more robust to these kinds of visual ambiguities. For supplementary videos with more qualitative results, see: \href{https://fsadeghi.github.io/Sim2RealViewInvariantServo}{https://fsadeghi.github.io/Sim2RealViewInvariantServo}

%% file: future.tex
In this paper, we described a learning-based approach to visual servoing, which can automatically and implicitly ``self-calibrate'' a robot in the process of performing a servoing task from a new viewpoint. Our method is based on training a deep convolutional recurrent neural network that can control a robot to reach user-specified query objects, implicitly learning to identify the effects of actions in image-space from the past history of observations and actions. The network is trained primarily in simulation, where supervised demonstrated data is easy to obtain automatically, and a novel adaptation procedure is used to adapt the visual layers of this model to the real world, using only a small number of labeled images.

The performance of our approach could be further improved in future work by incorporating depth or stereo perception, as well as by enabling the method to be finetuned to specific viewpoints at test time. Another exciting direction to explore in future work is how more complex manipulation skills can be performed from any viewpoint, using a similar approach. While we use reinforcement learning in our approach, our focus is on learning a viewpoint adaptation, rather than exploring pure reinforcement learning. We show how to effectively leverage simulation to solve such challenging task in simulation and adapting it to real world. Pure reinforcement learning supervision for learning view invariant manipulation skill can be explored in future work